\documentclass[runningheads]{llncs}
\usepackage[T1]{fontenc}
\usepackage{graphicx}
\usepackage{makecell}
\usepackage{bbding}
 \usepackage{amsmath}
 \usepackage{amssymb}
 \usepackage{ulem}
 \usepackage{algorithm}
\usepackage{algorithmic}

\usepackage[marginal]{footmisc}

\begin{document}
%
\title{DSDE: Using Proportion Estimation to Improve Model Selection for Out-of-Distribution Detection}
\titlerunning{DOS-Storey-based Detector Ensemble}
%
\author{Jingyao Geng\inst{1}\orcidID{0009-0008-2828-2843} \and
Yuan Zhang\inst{1}\orcidID{0009-0007-6309-7910} \and
Jiaqi Huang\inst{1}\orcidID{0009-0002-4611-0056}
\and 
Feng Xue\inst{2}\orcidID{0009-0007-2240-4283}
\and
Falong Tan\inst{3}\orcidID{0000-0003-4924-5378}
\and Chuanlong Xie\inst{1}\orcidID{0000-0003-4292-8782}
\and Shumei Zhang\inst{1}\orcidID{0000-0003-3840-2526}
}
\authorrunning{Geng et al. (2024)}

%
\institute{Beijing Normal University, China\\ 
\and
Shanghai Jiao Tong University, China\\
\and
Hunan University, China
}

\maketitle              
%

\footnote{\mailname{\, Chuanlong Xie, clxie@bnu.edu.cn}}

\begin{abstract}

Model library is an effective tool for improving the performance of single-model Out-of-Distribution (OoD) detector, mainly through model selection and detector fusion. 
However, existing methods in the literature do not provide uncertainty quantification for model selection results. Additionally, the model ensemble process primarily focuses on controlling the True Positive Rate (TPR) while neglecting the False Positive Rate (FPR).
In this paper, we emphasize the significance of the proportion of models in the library that identify the test sample as an OoD sample. This proportion holds crucial information and directly influences the error rate of OoD detection.
To address this, we propose inverting the commonly-used sequential p-value strategies. We define the rejection region initially and then estimate the error rate.
Furthermore, we introduce a novel perspective from change-point detection and propose an approach for proportion estimation with automatic hyperparameter selection.
We name the proposed approach as {\bf D}OS-{\bf S}torey-based {\bf D}etector {\bf E}nsemble (DSDE). 
Experimental results on CIFAR10 and CIFAR100 demonstrate the effectiveness of our approach in tackling OoD detection challenges. Specifically, the CIFAR10 experiments show that DSDE reduces the FPR from $11.07\%$ to $3.31\%$ compared to the top-performing single-model detector.

\end{abstract}
\section{Introduction}

\begin{figure}
\includegraphics[width=0.95\textwidth]{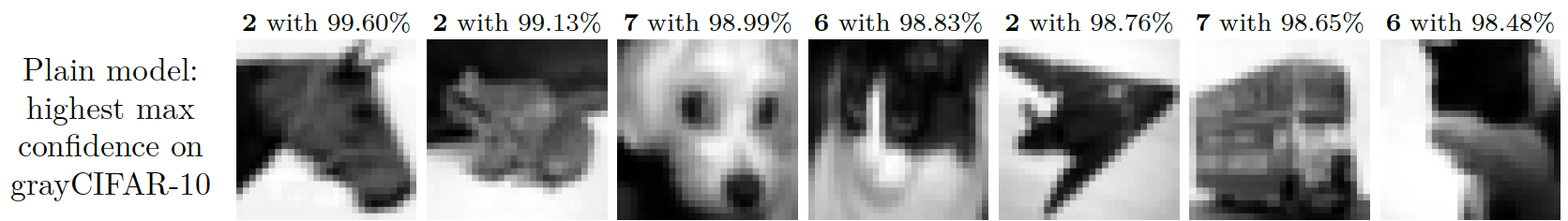}
\caption{Example of high-confidence predictions on images from CIFAR10.} \label{minist-cifar10}
\end{figure}

Deep neural networks (DNNs) demonstrate significant fitting capabilities.
Generally, a DNN's accuracy and robustness can be improved with an adequate number of training samples.
However, this also introduces new risks, such as well-trained DNNs producing highly confident predictions on unfamiliar samples.
For example, when a neural network classifier trained on the CIFAR-10 dataset \cite{cifar-10} is evaluated on the SVHN dataset \cite{hein2019relu}, it can confidently misclassify number plates as dogs, birds, and airplanes with nearly $100\%$ certainty.
Likewise, a model trained on the MNIST dataset \cite{mnist} can make predictions with exceptionally high confidence when presented with the grayscale version of the CIFAR-10 dataset.
Figure \ref{minist-cifar10} illustrates how the model incorrectly identifies a dog, cat, airplane, and truck as the numbers 7, 6, 2, and 7, respectively, with a confidence level of at least $98\%$.
These instances highlight the unreliability of deep neural networks, as they generate highly confident predictions for samples that were not encountered during training.

Most deep neural network models are trained in a closed-world environment, assuming that test data follows the same distribution as training data \cite{liang2017enhancing}. 
This assumption necessitates these models to generate predictions for all inputs.
However, a dependable classifier should refrain from making uninformed predictions on Out-of-Distribution (OoD) data as it impacts the model's reliability and generalization. This emphasizes the significance of OoD detection \cite{nguyen2015deep}. 
For instance, in autonomous driving, if a model fails to handle unforeseen events, it can result in accidents and put the driver at risk. Incorporating OoD detection enables the system to notify the driver and transfer control, mitigating the risk of accidents.
Likewise, in medical imaging, deep learning models aid physicians in detecting anomalies. If these models fail to accurately identify abnormal regions or assess new cases, it can lead to misdiagnoses and delayed treatments, thereby complicating patient care and placing a burden on healthcare facilities.

In recent years, research on OoD detection techniques has predominantly focused on single-model methods, with limited exploration of multi-model perspectives \cite{ood_survey1,ood_survey2,ood_survey3,ood_survey4}. 
Morningstar et al. enhanced OoD detection performance by combining multiple test statistics derived from generative models \cite{morningstar2021density}.
Haroush et al. employed Simes' and Fisher's methods to combine p-values calculated for each channel and layer of a deep neural network \cite{haroush2021statistical}.
Bergamin et al. showed that combining different types of test statistics using Fisher's method improves the accuracy of OoD detection \cite{bergamin2022model}.
Magesh et al. introduced a multi-detection framework for OoD detection, inspired by multiple testing, that combines statistics from neural network models \cite{magesh2022multiple}.
Xue et al. proposed the ZODE algorithm, which integrates multiple pre-trained models from a model zoo and selects those capable of identifying test inputs as OoD samples \cite{xue2022boosting}. The ZODE method standardizes multiple detection scores into p-values and employs the Benjamini $\&$ Hochberg (BH) method \cite{benjamini1995controlling} for OoD decision-making.

Detecting OoD samples using multiple pre-trained models presents more complex challenges compared to single-model OoD detection, i.e. controlling True Positive Rate (TPR).
While controlling the TPR is typically done for a single-model detector, managing the combined TPR rate becomes necessary when employing multiple OoD detectors.
Generally, controlling TPR is derived from the upper bound of the false discovery rate (FDR) using intricate sequential p-value rejection methods based on observed data \cite{magesh2022multiple,xue2022boosting}. 
A sequential p-value method produces an estimate $\hat{k}$, where the detectors associated with $p_{(1)},p_{(1)},\cdots,p_{(\hat{k})}$ identify the test input as an OoD sample. Here, $p_{(1)} \leq p_{(2)} \leq \cdots \leq p_{(m)}$ represent the ordered observed p-values.
Nevertheless, this procedure exhibits several significant weaknesses. Firstly, $\hat{k}$ is a random variable without a measure of its uncertainty. Ideally, $\text{TPR} \geq 1-\alpha$, but the reliability of these methods varies on a case-by-case basis. 
Another weakness is that the TPR is controlled concurrently for all values of $m_0$ (the number of pre-trained models identifying the test input as an ID sample), without utilizing any information about $m_0$ in the observed p-values. Incorporating this information can result in a less stringent and more sensitive procedure for OoD data while maintaining robust TPR control.

This paper aims to employ conventional and straightforward statistical concepts to control the TPR.
While a sequential p-value method fixes the error rate and estimates the corresponding rejection region, we propose the reverse approach: fixing the rejection region and estimating its associated error rate.
We incorporate the Storey correction method \cite{storey2002direct} into OoD Detection and develop a novel algorithm to leverage multiple pre-trained models.
For each test input, we empirically estimate the number of pre-trained models that classify the test input as an ID sample. We utilize this information to enhance the OoD detector.

However, hyperparameter tuning remains a challenging task, lacking consensus on the optimal approach~\cite{jiang2008estimating}.
Like the existing sequential p-value approaches \cite{magesh2022multiple,xue2022boosting}, our proposed approach also faces challenges of high variance in model selection results.
To mitigate this issue, \cite{kostic2023change} introduce a hyperparameter selection approach named 'DOS-Storey' that leverages insights from change-point detection. This method approximates the p-value plot using a piecewise linear function with a single change-point, enabling the selection of the hyperparameter based on the p-value at the change-point location.
The DOS-Storey estimator automatically selects the optimal hyperparameter value and demonstrates reduced bias and variance, thereby enhancing performance stability.
Consequently, we employ the DOS-Storey estimator to determine the number of pre-trained models that identify the test input as an OoD sample.
We name the proposed approach as {\bf D}OS-{\bf S}torey-based {\bf D}etector {\bf E}nsemble (DSDE).

The primary contributions and findings of this study are outlined below:

1. Contrary to the sequential p-value method, where the error rate is determined to infer its corresponding rejection region, our approach fixes the rejection region first and then estimates the error rate. We utilize a model library to determine the proportion of models capable of classifying the test input as OoD samples. By employing this approach, we mitigate the issue of missed detections for OoD samples and improve detection accuracy.

2. We leverage the insight from change-point detection and utilize the DOS-Storey estimator to select optimal hyperparameter values. Unlike the Storey proportion estimator method, the DOS-Storey estimator autonomously selects $\lambda$ values, resulting in lower bias levels and reduced variances. This characteristic ensures a more stable performance.

3. In this study, we evaluate the effectiveness of the proposed DSDE detector using comprehensive experiments.
On the CIFAR10 dataset, the experimental results show that the proposed algorithm reduces the average FPR from $11.07\%$ to $3.31\%$ compared to the top-performing single-model detector. Similarly, on CIFAR100, a reduction in FPR from $48.75\%$ to $41.28\%$ is achieved.
Furthermore, this algorithm can be integrated with various baseline methods for post-detection, and the DSDE-KNN detector exhibits superior performance.
On the CIFAR10 dataset, the DSDE-KNN detector achieves a significant reduction in FPR from $20.74\%$ to $3.31\%$ compared to the top-performing baseline method. Similarly, on CIFAR100, the enhanced detector achieves a reduction in FPR from $64.14\%$ to $41.28\%$.

\section{Background}

\subsection{Out-of-Distribution detection}

Denote the training inputs as $D_{in} = \{x_{i}\}_{i=1}^{n}$ and $P_{in}$ representing the training distribution of $x$. Let $\phi$ be a pre-trained feature extractor or classifier. We write an OoD detector as
\begin{equation}\label{eq1}
G(x^{*}, \phi) = \left\{ \begin{array}{rcl}
                & \text{ID}, & \mbox{if} \,\, S(x^{*},\phi) > \lambda_\phi;\\
                & \text{OoD},  & \mbox{if} \,\, S(x^{*},\phi) \leq \lambda_\phi;
                \end{array}\right.
\end{equation}
where 
$S(\cdot, \cdot)$ is a scoring function depends on the test input $x^*$ and the model $\phi$.
The threshold $\lambda_\phi$ is determined by controlling the TPR level at $1-\alpha$, i.e.,
\begin{equation}\label{eq2}
    \lambda_\phi = \inf_{s \in \mathbb{R}} \{ s: \hat F(s;\phi) \geq \alpha \} \quad \text{with} \quad \hat F(s; \phi) = \frac{1}{n} \sum_{i=1}^n \mathbb{I} \big\{ S(x_i,\phi) \leq s \big\}.
\end{equation}
Here $\mathbb{I}\{\cdot\}$ represents the indicator function. 
Without access to the OoD samples, the OoD detection task can be reformulated as a one-sample hypothesis testing problem:
\begin{equation}\label{eq3}
H_{0}:x^{*}\sim P_{in} \quad \text{v.s.}\quad H_{1}: x^{*}\sim Q \,\, \text{with}\,\, Q \neq P_{in}.
\end{equation}
Let $s^{*} = S(x^{*},\phi)$ denote the score of the test input $x^*$. Then the p-value of $x^*$ is given by
\begin{equation}\label{eq4}
p^* = P( S(x,\phi) <s^{*}|x \sim P_{in}) \approx \hat F(s^*;\phi).
\end{equation}

\subsection{Using model library}

Xue et al. proposed an OoD detection algorithm called ZODE~\cite{xue2022boosting}. This method integrates multiple pre-trained models into a model zoo. Given any test input, each pre-trained model in the model zoo performs detection based on the same scoring function and provides a score. 
Experiments show that the ZODE method is suitable for various OoD detection tasks, effectively controlling the TPR to reach the target level while maintaining a low FPR. The limited feature extraction capability of a single trained model leads to insufficient generalization ability. In contrast, the ZODE method leverages multiple pre-trained models to capture a wider variety of features, resulting in more robust OoD detection.

To enhance the suitability of the detection decision statistics of the OoD detector for multiple comparison corrections, this study converts the detection score $S(x^{*},\phi_{i})$ into a p-value. Furthermore, as the 
range of the detection score $S(x^{*},\phi_{i})$ vary with the pre-trained model $\phi_{i}$, the p-value serves to standardize the detection decision statistics of multiple single-model detectors. Additionally, when the OoD detector bases its decisions on p-values, the outcomes hold statistical significance.

When provided with a test sample $x^{*}$ and assuming $x^{*}$ receives a detection score $s_{i}^{*} = S(x^{*},\phi_{i})$ from a single-model OoD detector $G(x,\phi_{i})$, the expression for calculating the p-value can be obtained based on the definition of the p-value as:
\begin{equation}\label{eq5}
p_{i} = P(s<s_{i}^{*}|x \sim P_{in}) = P(S(x,\phi_{i})<s_{i}^{*}|x \sim P_{in}) = F(s_{i}^{*},\phi_{i}).
\end{equation}
In Eq.~(\ref{eq5}) above, $p_{i}$ denotes the p-value of $x^{*}$ under a single-model detector $G(x,\phi_{i})$, while $F(\cdot,\phi_{i})$ represents the cumulative distribution function of the detection scores $S(x,\phi_{i})$ for in-distribution samples under $G(x,\phi_{i})$.

If the $p_{i}$ is less than $\alpha$, $G(x,\phi_{i})$ categorizes $x$ as an OoD sample. When a test input $x^{*}$ is concurrently detected by an ensemble detector $G(x^{*},M)$ comprising $m$ individual OoD detectors, a sequence of p-value $p_{seq} = \{p_{1},p_{2},...,p_{m}\}$ related to $x^{*}$ will be generated.Converting the detection scores $S(x^{*},\phi_{i})$ into p-values does not enable a direct comparison of the p-value sequence $p_{seq}$ with $\alpha$. Such a comparison would result in an inflation of the Type I error rate.
To mitigate this concern, a multiple testing procedure can be utilized to adjust the p-value threshold.

\subsection{Storey FDR correction method}

Obtaining the true cumulative distribution function $F(\cdot,\phi_{i})$ is challenging. In cases where the samples in the validation set are independently and identically distributed, the empirical cumulative distribution function $\hat{F}(s;\phi_{i})$ of the validation set samples' scores $S(x,\phi_{i})$ can serve as an approximation to the true cumulative distribution function. 

\begin{equation}\label{eq6}
\hat{F}(s;\phi_{i}) = \frac{1}{n} \sum_{j=1}^{n}\mathbb{I}\{S(x_{j},\phi_{i}) \leq s \}.
\end{equation}

Eq.~(\ref{eq5}) can be elaborated as follows.

\begin{equation}\label{eq7}
p_{i} =  P(S(x,\phi_{i})<s_{i}^{*}|x \sim P_{in}) = \hat{F}(s_{i}^{*},\phi_{i}).
\end{equation}

Storey et al. \cite{storey2003positive} introduced the Storey FDR correction method, which computes the q-value for every p-value to regulate the False Discovery Rate. The detailed testing procedure is outlined as follows:

1.Initially, organize the sequence of p-values $p_{seq} = (p_{1},p_{2},...,p_{m})$ in ascending order to derive $p_{(1)}<p_{(2)}<,...,<p_{(m)}$.

2.Calculate the estimated values of $\hat{\pi}_{0}(\lambda)$ for $\lambda = 0, 0.01, 0.02, ..., 0.95$.

\begin{equation}\label{eq8}
\hat{\pi}_{0}(\lambda) = \frac{\# \{ p_{j} > \lambda \}}{m(1-\lambda)}. 
\end{equation}

3.Calculate the q-value for $i = 1, 2, ..., m$:
\begin{equation}\label{eq9}
\hat{q}(p_{(i)}) = \min_{\alpha \leq p_{j}} 
\frac{\hat{\pi}_{0} \alpha m}{\# \{p_{j} \leq \alpha\} } = \frac{\hat{\pi}_{0} m p_{(i)}}{i}.
\end{equation}

4.Compare $\hat{q}(p_{(i)})$ with the provided significance level $\alpha$ and calculate the index $\hat{k}$ corresponding to the largest $\hat{q}(p_{(i)})$ value that is less than $\alpha$:
\begin{equation}\label{eq10}
\hat{k} = \arg\max \{k:\hat{q}(p_{(i)}) \leq \alpha \}.
\end{equation}
Subsequently, reject the null hypothesis associated with $p_{(1)},p_{(2)},...,p_{(\hat{k})}$ based on the computed index $\hat{k}$ value from Eq.~(\ref{eq10}); if such $\hat{k}$ does not exist, retain all null hypotheses. Q-values, by definition, represent False Discovery Rate (FDR) values. To maintain the FDR below 0.05, set the $\alpha$ value to 0.05; for FDR control below 0.1, set $\alpha$ to 0.1. The Storey FDR correction method efficiently manages the FDR using q-values.

\section{Out-of-Distribution Detection based on Model Library}

\subsection{Ratio estimator}
Storey et al.~\cite{storey2002direct} modeled the distribution of p-values as a mixture with a cumulative distribution function:
\begin{equation}\label{eq11}
F(x) = \pi_{0}x + \pi_{1}F_{1}(x),~ x \in [0,1],
\end{equation}
where $\pi_{0}$ represent the proportion of true null hypotheses among all tests and $\pi_{1} = 1-\pi_{0}$. 
Storey introduced a set of proportional interpolation estimators for $\pi_{0}$~\cite{storey2002direct}, as depicted in the equation below:
\begin{equation}\label{eq12}
\hat{\pi}_{0}(\lambda) = \frac{1-\hat{F_{n}}(\lambda)}{1-\lambda}, ~ \lambda \in (0,1)
\end{equation}
The empirical distribution function $\hat{F_{n}}(\cdot)$ is derived from p-value observations, while $\lambda$ serves as a tunable parameter within the range $(0,1)$. Typically, a smaller $\lambda$ leads to increased bias, whereas choosing a value closer to 1 for $\lambda$ raises the variance of the ratio estimator~\cite{broberg2005comparative}. Hence, Storey suggests setting $\lambda$ at 0.5.

\subsubsection{Change-Point Points in the DOS-Storey Ratio Estimator.}


The emergence of change-point points is linked to the linear deviation of p-values. Anica Kostic and colleagues employed p-value information to compute the estimated value $\hat{k}$ for the change-point point and introduced a sequence statistic called DOS (Difference of Slopes)~\cite{jiang2008estimating}, as illustrated in the following equation:
\begin{equation}\label{eq13}
d_{\beta}(i) =\frac{p_{(2i)}-p_{(i)}}{i^{\beta}} - \frac{p_{(i)}}{i^{\beta}} 
 = \frac{p_{(2i)}-2p_{(i)}}{i^{\beta}}.
\end{equation}
The parameter $\beta$ is adjustable, where $\beta \in [1/2, 1]$. Varying values of $\beta$ result in diverse interpretations of the sequence statistic DOS. By maximizing the DOS statistic, the estimated value $\hat{k}$ for the change-point point can be identified. The estimation process for the change-point point is outlined as follows:

1.Initially, arrange the p-values of $m$ independent tests in ascending order: $p_{(1)}<p_{(2)}<...<p_{(m)}$.

2.Calculate the sequence statistic DOS for $i=1,2,...,m/2$ using Eq.~(\ref{eq13}).

3.The index corresponding to the maximum value in the DOS sequence statistic provides the estimated position of the change-point.
\begin{equation}\label{eq14}
\hat{k_{\beta}} = \mathop{\arg\max}\limits_{m*c_{m} \leq i \leq n/2}d_{\beta}(i).
\end{equation}
To maintain the asymptotic properties of the proportion estimator $\hat{\pi}_{0}^{\beta}(\lambda)$ for the non-random sequence selection $c_{m}$, Anica Kostic and colleagues recommend selecting $c_{m} = \epsilon \in (0,1)$~\cite{jiang2008estimating}.

\subsubsection{Dos-Storey True Zero Proportion Estimator.}

In the context of estimating $\hat{\pi}_{0}(\lambda)$, Anica Kostic and colleagues introduced the DOS-Storey proportion estimator~\cite{kostic2023change}, which involves determining the value of $\hat{k_{\beta}}$ at the change-point. The estimation process entails inserting the p-value associated with $\hat{k_{\beta}}$ into Eq.~(\ref{eq12}) to compute the DOS-Storey true zero proportion estimator, denoted as $\hat{\pi}_{0}(\lambda)$, according to the following formula:
\begin{equation}\label{eq15}
\hat{\pi}_{0}^{\beta} = (1- \frac{\hat{k}_{\beta}}{n})(1-p_{(\hat{k}_{\beta})}).
\end{equation}
The simulation results conducted by the author reveal the superiority of this estimator over the Storey method's proportion estimator $(\lambda = 0.5)$ across various parameter settings ($\beta$ = 1/2 or $\beta$ = 1), exhibiting reduced bias and variance. Specifically, the simulation outcomes highlight that the DOS-Storey estimator exhibits the lowest Root Mean Square Error (RMSE)~\cite{kostic2023change}, underscoring its consistent performance and robustness against data interference. Consequently, the paper will introduce the DOS-Storey estimator into the Storey FDR correction method.

\subsection{Algorithm design}

\begin{figure}[h]
\begin{center}
\includegraphics[width=0.95\textwidth]{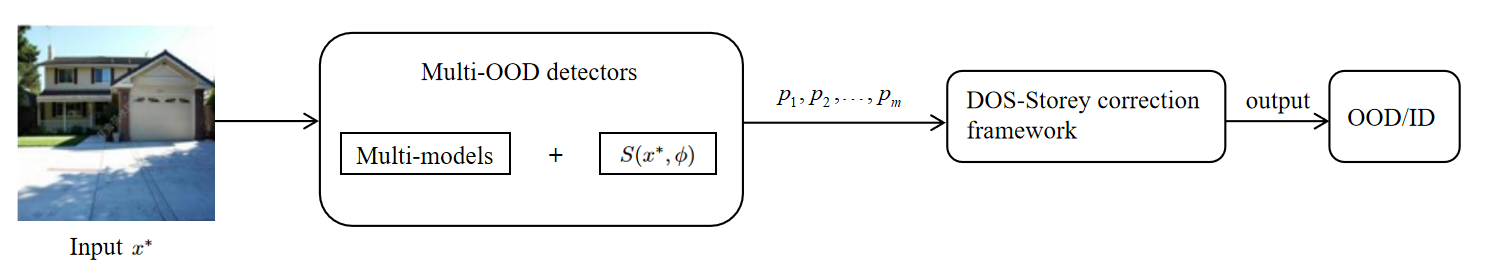}
\end{center}
\caption{Flowchart of DSDE.} \label{flowchart}
\end{figure}

The DOS-Storey correction framework is expanded to incorporate model-based OoD detection, resulting in the development of a model-based OoD detection. Algorithm \ref{algorithm1} offers a detailed description of the algorithm steps,and Figure \ref{flowchart} presents a simple flowchart of the Algorithm\ref{algorithm1}.

\begin{algorithm}[t]
\caption{{DSDE}: 
DOS-Storey-based Detection Ensemble
}\label{algorithm1}
\begin{algorithmic}[1]
\REQUIRE model library $M = \{\phi_{1},\phi_{2},...,\phi_{m}\}$, 
scoring function $S(x,\phi)$, significance level $\alpha$, 
hyperparameters $\beta$, $c_{m}$, validation data $\{x_i\}_{i=1}^n$, test input $x^{*}$\\ 
\FOR {$1 \leq j \leq m$} 
\STATE Compute $s^*_j = S(x^*; \phi_j)$ and $p_{j}^{*} = \hat{F}(s_{j}^{*};\phi_{j})$ according to Eq.~(\ref{eq4});
\ENDFOR\\
\STATE Sort $\{p_1, \ldots, p_m\}$  in
ascending order: $\{p_{(1)}, \ldots, p_{(m)}\}$;
\STATE Search $\hat{k_{\beta}}$ according to Eq.~(\ref{eq14});
\STATE compute $\hat{\pi}_{0}$ according to Eq.~(\ref{eq15});
\STATE Search $k$ according to Eq.~(\ref{eq10});
\IF { $k$ does not exists,}  \STATE \textbf{output:} $x^*$ is an ID sample;
\ELSE 
\STATE \textbf{output:} the $k$ pre-trained models corresponding to the $p$-values $p_{(1)} \cdots p_{(k)}$;
\STATE \textbf{output:} $x^*$ is an OOD sample.
\ENDIF
\end{algorithmic}
\end{algorithm}

The detection process of $G(x^{*},M)$ on $x^{*}$ is divided into two stages. Initially, with a model repository $M = \{ \phi_{1},...,\phi_{m} \}$ containing $m$ pre-trained models and a scoring function $S(x,\phi)$, these conditions establish a model-based outlier detector $G(x^{*},M) = \{ G(x^{*},\phi_{j}), \forall \phi_{j} \in M \}$. In the first stage, $G(x^{*},M)$ generates a detection statistic p-value as output, which is then utilized as input for the second stage. The second stage involves adjusting the p-value threshold using the DOS-Storey correction method to ascertain if the input $x^{*}$ is an outlier sample.

The DSDE algorithm's core resides in employing the DOS-Storey correction framework to systematically integrate the detection decision statistics from multiple detectors. This method introduces an estimate of $\pi_{0}$, enabling the identification of a greater number of outlier samples; however, it may adversely affect the detection accuracy of in-distribution samples, specifically impacting the True Positive Rate (TPR). When the estimate $\hat{\pi}_{0}$ closely approximates the true proportion $\pi_{0}$ during the detection of in-distribution samples, the algorithm's effect on their detection accuracy is minimal. This paper demonstrates in Section 4, through empirical experiments, that the DSDE algorithm can maintain the TPR close to the target level, thereby ensuring accurate detection of in-distribution samples while simultaneously identifying more OoD samples.

\section{Experiment}

In this section, we conducts comprehensive experiments on widely used benchmarks for OoD detection tasks to demonstrate that the proposed approach offers increased applicability, accuracy, and sensitivity. 

\subsection{Implementation details}

We employ CIFAR10~\cite{alex2009learning} and CIFAR100~\cite{alex2009learning} datasets for in-distribution datasets and utilize five OoD datasets, SVHN~\cite{netzer2011reading}, LSUN~\cite{yu2015lsun}, iSUN~\cite{xu2015turkergaze}, Texture~\cite{cimpoi2014describing}, and Places365~\cite{zhou2017places} to assess OoD detection performance.
We consider four baseline OoD detection methods, including MSP~\cite{hendrycks2016baseline}, ODIN~\cite{liang2017enhancing}, Energy~\cite{liu2020energy}, and KNN method~\cite{sun2022out}. 
Three evaluation metrics are employed to assess the detection performance. These metrics comprise the True Positive Rate (TPR), the False Positive Rate (FPR) when the TPR is 95\%, and the Area Under the Receiver Operating Characteristic Curve (AUC).
Here the positive samples are the training data and the negative samples refer to the OoD datasets.

To fully showcase the advantages and effectiveness of the DSDE algorithm, our model library consists of seven models, both widely employed in OoD detection tasks. The choice of pre-trained models is approached from various angles: firstly, the consideration of network depth impact on feature extraction entails the inclusion of ResNet models \cite{he2016deep} of varying depths - ResNet18, ResNet34, ResNet50, ResNet101, and ResNet152. Secondly, the impact of models with distinct structures on feature extraction is examined, with the addition of a 100-layer DenseNet~\cite{huang2017densely} to the model library. Furthermore, this study analyzes the effect of loss functions during model training. The cross-entropy loss function is utilized to train all aforementioned models, while the SupConLoss~\cite{khosla2020supervised} function (Supervised Contrastive Loss) is employed for training ResNet18*, referencing the experimental setup of ~\cite{sun2022out}. Overall, the model library illustrates sufficient diversity in architecture and training objectives.

\subsection{Experiments on CIFAR10}

In this section, we take KNN method~\cite{sun2022out}. as the scoring function, with the hyperparameter $k$ set to 50. To achieve the TPR level closing to 95\%, the parameters in Algorithm \ref{algorithm1} are assigned values $c_{m} = 2/7$ and $\beta=1$.

\begin{table}[t]
\caption{The results of the single-model detector and our proposed {DSDE} detector on CIFAR10.}\label{tab1}
\resizebox{\textwidth}{!}{
\begin{tabular}{|l|l|l|l|l|l|l|l|}
\hline
 \makecell[c]{Model/ (\%)} & \makecell[c]{TPR \\ ~}  & \makecell[c]{SVHN \\(FPR/AUC)} & \makecell[c]{LSUN \\ (FPR/AUC)} & \makecell[c]{iSUN \\ (FPR/AUC)} & \makecell[c]{DTD \\ (FPR/AUC)} & \makecell[c]{Places365 \\ (FPR/AUC)} &
 \makecell[c]{Average \\ (FPR/AUC)}\\
\hline
ResNet18* & 95.00 & 2.43/{\bf 99.52} & 1.78/99.48 & 20.06/96.74 & 8.09/98.57 & 22.96/95.32 & 11.06/ 97.93\\
ResNet18 & 95.00 & 27.95/95.49 & 18.48/96.84 & 24.65/95.52 & 26.70/94.97 & 48.19/90.01 & 29.19 /94.57\\
ResNet34 & 95.00 & 26.47/95.85 & 10.20/98.39 & 29.41/95.16 & 31.61/94.53 & 36.47/92.74& 26.83/95.34\\
ResNet50 & 95.00 & 17.31/97.40 & 7.08/98.83 & 17.31/97.26 & 20.82/96.59 & 41.51/91.60 & 20.81/96.34\\
ResNet101 & 95.00 & 25.75/96.12 & 6.66/98.90 & 19.85/96.80 & 18.42/96.89 & 40.73/92.15 & 22.28/96.17\\
ResNet152 & 95.00 & 34.98/94.98 & 7.26/98.88 & 22.32/96.66 & 20.78/96.60 & 38.61/92.37 & 24.79/95.90\\
DenseNet & 95.00 & 10.31/98.18 & 7.88/98.60 & 10.86/97.94 & 20.85/96.25 & 50.08/88.92 & 20.00/95.98 \\
\hline
DSDE & 94.91 & {\bf 1.92}/99.46 & {\bf 1.37/99.63} & {\bf 4.89/98.77} & {\bf 0.12/99.88} & {\bf 8.26/98.17} & {\bf 3.31/99.18}\\
\hline
\end{tabular}}
\end{table}

Table~\ref{tab1} illustrates the comparison between the single-model-base detectors and the {DSDE} detector. 
We can find that the OoD detector using the model library minimally affects TPR retention at the designated threshold. 
The TPR of the DSDE detector falls slightly below $95\%$, exhibiting a mere $0.09\%$ deviation from the target threshold. This suggests that the estimator's assessment of $\hat{\pi}_{0}(\lambda)$ minimally affects TPR during ID sample detection, thereby affirming the efficacy of the methodology. Upon longitudinal comparison of Table~\ref{tab1}, the DSDE detector exploits the complementary aspects of individual model detectors. Our proposed method yields a reduced FPR compared to any single-model detectors demonstrating the capability of leveraging model diversity to detect the OoD samples misclassified by single-model detectors.
Comparing to the best-performance of single-model detectors, our method decreases the average FPR from $11.06\%$ to $3.31\%$, representing a notable $70.07\%$ relative improvement. These findings illustrate that the effectiveness of the proposed detector is not solely dependent on any single-model detector. Our approach successfully merges single-model detectors, resulting in enhanced OoD detection capabilities.

\begin{table}[t]
\caption{Comaprison with Baseline Methods on CIFAR10.}\label{tab2}
\resizebox{\textwidth}{!}{
\begin{tabular}{|l|l|l|l|l|l|l|l|}
\hline
 \makecell[c]{Method/(\%)} & \makecell[c]{TPR \\ ~}  & \makecell[c]{SVHN \\(FPR/AUC)} & \makecell[c]{LSUN \\ (FPR/AUC)} & \makecell[c]{iSUN \\ (FPR/AUC)} & \makecell[c]{DTD \\ (FPR/AUC)} & \makecell[c]{Places365 \\ (FPR/AUC)} & \makecell[c]{Average \\ (FPR/AUC)}\\
\hline
MSP & 95.00 & 59.88/90.48 & 28.57/95.99 & 45.44/93.54 & 57.14/90.14 & 58.32/89.24 & 49.87/91.88\\
DSDE-MSP & 94.96 & 49.76/93.04 & 13.96/97.68 & 28.94/95.74 & 40.20/94.89 & 41.42/94.74 & 34.86/95.22\\
\hline
ODIN & 95.00 & 53.36/86.83 & 18.39/93.07 & 32.18/89.29 & 56.96/83.54 & 49.18/84.59 & 42.02/87.46\\
DSDE-ODIN & 94.92 & 40.77/91.89 & 0.92/{\bf 99.65} & 3.44/{\bf 99.08} & 14.02/97.55 & 17.96/96.99 & 15.42/97.03\\
\hline
Energy & 95.00 & 54.49/87.60 & 4.09/99.07 & 11.09/97.81 & 50.17/85.96 & 48.56/88.27 & 33.68/91.74\\
DSDE-Energy & 94.88 & 45.73/92.60 & 2.78/99.33 & 13.88/97.26 & 32.54/95.40 & 16.63/97.19 & 22.31/96.36\\
\hline
KNN & 95.00 & 20.74/96.79 & 8.48/98.56 & 20.64/96.58 & 21.04/96.34 & 39.79/91.87 & 20.74/96.79\\
DSDE-KNN & 94.91 & {\bf 1.92/99.46} & {\bf 1.37}/99.63 & {\bf 4.89}/98.77 & {\bf 0.12/99.88} & {\bf 8.26/98.17} & {\bf 3.31/99.18}\\
\hline
\end{tabular}}
\end{table}

Table~\ref{tab2} show that our approach is compatible with several baseline methods. Observations of the TPR of different detectors indicate that DSDE with various baseline methods had an insignificant impact on the target TPR level of $95\%$, with values predominantly around $94.90\%$. Evaluation of the FPR values across various detection methods reveals that the DSDE detector displays considerably lower FPR values than the corresponding baseline methods, accompanied by a notable rise in the AUC value. This underscores the potential of our approach to enhance the OoD detection performance of baseline methods.
According to Table~\ref{tab2}, we can also find that combining DSDE with the KNN score resulted in superior performance, where KNN emerged as the most effective baseline method. The DSDE-KNN detector demonstrated a remarkable reduction in the FPR from $20.74\%$ to $3.31\%$, reflecting an impressive $84.04\%$ relative improvement.

\subsection{Experiments on CIFAR100}

\begin{table}[t]
\caption{The results of the single model detector and our proposed DSDE detector on CIFAR100.}\label{tab3}
\resizebox{\textwidth}{!}{
\begin{tabular}{|l|l|l|l|l|l|l|l|}
\hline
 \makecell[c]{Model/(\%)} & \makecell[c]{TPR \\ ~}  & \makecell[c]{SVHN \\(FPR/AUC)} & \makecell[c]{LSUN \\ (FPR/AUC)} & \makecell[c]{iSUN \\ (FPR/AUC)} & \makecell[c]{DTD \\ (FPR/AUC)} & \makecell[c]{Places365 \\ (FPR/AUC)} & \makecell[c]{Average \\ (FPR/AUC)} \\
\hline
ResNet18 & 95.00 & 63.99/86.03 & 71.23/78.44 & 70.46/84.19 & 63.33/83.84 & 79.24/76.31 & 69.65/81.76\\
ResNet34 & 95.00 & 62.49/87.81 & 71.53/81.98 & 70.68/83.83 & 74.24/79.60 & 79.53/76.77 & 71.69/82.00 \\
ResNet50 & 95.00 & 56.26/87.63 & 67.04/{\bf 82.65} & 56.78/87.50 & 55.14/86.69 & 77.30/76.67 & 62.50/84.23\\
ResNet101 & 95.00 & 52.53/89.72 & 76.04/78.00 & 63.15/86.75 & 61.54/85.05 & 79.11/77.14 & 66.47/83.33\\
ResNet152 & 95.00 & 54.83/86.06 & 75.78/78.00 & 57.49/87.84 & 62.89/85.18 & 77.89/77.20 & 65.78/82.86\\
DenseNet & 95.00 & 39.68/89.99 & 53.46/83.76 & 37.55/{\bf 92.50} & 26.12/93.94 & 86.96/69.60 & 48.75/85.96\\
\hline
DSDE & 94.62 & {\bf 34.21/90.29} & {\bf 52.22}/81.87 & {\bf 37.18}/90.90 & {\bf 14.36/97.39} & {\bf 68.44/85.64} & {\bf 41.28/89.22}\\
\hline
\end{tabular}}
\end{table}

The experimental procedure followed a consistent approach with CIFAR10, with the TPR measured on the CIFAR100 test set. Table~\ref{tab3} illustrates results from various OoD detectors on the same dataset. The conclusions align with those on CIFAR10, indicating that model library-based OoD detectors maintain low FPR values with minimal impact on TPR. Notably, the proposed method in Table~\ref{tab3} shows lower FPR values than individual detectors across diverse OoD detection tasks, based on the same baseline method, with varying improvements in AUROC values. This suggests the proposed method can better detect OoD samples by leveraging differences in detection from various models. Comparing to the best-performance of single-model detectors, our method decreases the average FPR from $48.75\%$ to $41.28\%$, representing a notable $15.32\%$ relative improvement.

\begin{table}
\caption{Comaprison with Baseline Methods on CIFAR100.}\label{tab4}
\resizebox{\textwidth}{!}{
\begin{tabular}{|l|l|l|l|l|l|l|l|}
\hline
 \makecell[c]{Method/(\%)} & \makecell[c]{TPR \\ ~}  & \makecell[c]{SVHN \\(FPR/AUC)} & \makecell[c]{LSUN \\ (FPR/AUC)} & \makecell[c]{iSUN \\ (FPR/AUC)} & \makecell[c]{DTD \\ (FPR/AUC)} & \makecell[c]{Places365 \\ (FPR/AUC)} & \makecell[c]{Average \\ (FPR/AUC)} \\
\hline
MSP & 95.00 & 77.29/80.07 & 77.83/80.74 & 78.08/79.44 & 82.33/75.92 & 81.25/76.35 & 79.36/76.69\\
ODIN & 95.00 & 49.44/{\bf 91.69} & 56.32/{\bf 88.10} & 92.20/65.89 & 78.18/73.62 & 88.23/64.96 & 72.87/76.85\\
Energy & 95.00 & 74.00/82.90 & 71.56/83.73 & 74.59/81.88 & 80.84/77.11 & 81.39/76.14 & 76.48/80.35\\
KNN & 95.00 & 54.96/87.87 & 69.18/80.47 & 59.35/87.10 & 57.21/85.72 & 80.01/75.62 & 64.14/83.37\\
DSDE-KNN & 94.96 & {\bf 34.21}/90.29 & {\bf 52.22}/81.87 & {\bf 37.18/90.90} & {\bf 14.36/97.39} & {\bf 68.44/85.64} & {\bf 41.28/89.22}\\
\hline
\end{tabular}}
\end{table}

Table~\ref{tab4} was generated through a vertical comparison of experimental results from various detection methods on the same OoD dataset, highlighting the superior performance of the DSDE-KNN detector in terms of FPR values and AUROC values across diverse datasets. Additionally, the DSDE-KNN detector demonstrates exceptional performance on challenging OoD datasets. For example, in the texture dataset DTD, where baseline methods have FPR values above 50\%, the DSDE-KNN detector reduces the FPR to 14.36\%, resulting in a remarkable 74.90\% increase in relative detection accuracy compared to the best baseline method.

\subsection{Comparison of Different Combination Schemes.}

This paper aims to develop a systematic approach for combining models that retains a reduced FPR while minimally impacting the TPR at the desired level. Comparative experiments are conducted in this section to highlight the successful implementation of Algorithm \ref{algorithm1} in achieving this research goal. CIFAR10 is utilized as the in-distribution dataset, aligning with the model library and OoD dataset configuration outlined in Section 4.2. The KNN method is chosen as the standalone detector for OoD detection, with a hyperparameter $k$ set to $50$. Additionally, parameters are specified as $c_{m} = 2/7$ and $\beta = 1$.

The combination strategies comprise naive decision-making(native), multiple testing, and ensemble methods. The multiple testing approach involves a comparison between Bonferroni, BY, and BH methods and the proposed DSDE method in this paper. In the ensemble scheme, the conventional voting method in classification tasks is employed, following the majority decision principle. Samples are classified as OoD if at least 50\% or 60\% of the model detectors in the library recognize them as such. The experimental outcomes and normalized results are presented in Table~\ref{tab6}.

\begin{table}
\caption{Performance Comparison of Different Combination Schemes.}\label{tab6}
\resizebox{\textwidth}{!}{
\begin{tabular}{|l|l|l|l|l|l|l|l|}
 \hline
  \makecell[c]{Method/(\%)} & \makecell[c]{TPR \\ ~}  & \makecell[c]{SVHN \\(FPR/AUC)} & \makecell[c]{LSUN \\ (FPR/AUC)} & \makecell[c]{iSUN \\ (FPR/AUC)} & \makecell[c]{DTD \\ (FPR/AUC)} & \makecell[c]{Places365 \\ (FPR/AUC)} & \makecell[c]{Average \\ (FPR/AUC)} \\
 \hline
 Naive & 69.87 & {\bf 0.16}/99.18 & {\bf 0.12}/99.39 & {\bf 0.53}/98.21 & {\bf 0.02}/99.72 & {\bf 0.10}/97.39 & {\bf 0.19}/98.78\\
 Bonferroni & 95.07 & 2.84/99.26 & 2.01/99.47 & 7.82/98.22 & 0.37/99.83 & 15.50/97.40 & 5.71/98.84\\
 BY & 98.11 & 6.96/99.42 & 4.64/99.61 & 15.71/98.69 & 1.38/99.87 & 34.03/98.02 & 12.54/99.12\\
 BH & 94.96 & 2.12/99.43 & 1.50/99.61 & 5.48/98.70 & 0.16/99.88 & 9.91/97.99 & 3.83/99.12\\
 Voting(50\%) & 99.98 & 11.99/99.92 & 4.29/99.98 & 12.72/99.93 & 3.28/{\bf 100.00} & 28.25/{\bf 99.93} & 12.11/99.95\\
 Voting(60\%) & 100.00 & 22.54/{\bf 99.96} & 7.68/{\bf 99.99} & 21.78/{\bf 99.96} & 15.28/{\bf 100.00} & 58.56/99.90 & 25.17/{\bf 99.96}\\
 DSDE-KNN & 94.91 & 1.92/99.46 & 1.37/99.63 & 4.89/98.77 & 0.12/99.88 & 8.26/98.17 & 3.31/99.18\\
 \hline
 \end{tabular}}
 \end{table}

Table~\ref{tab6} illustrates the ability of the naive approach to consistently keep FPR at the lowest levels across various OoD datasets. However, it falls short in regulating TPR near the target level, displaying significantly lower TPR levels than intended, highlighting its unsuitability for in-distribution sample detection and increased risk of Type I errors. The voting schemes, including voting (50\%) and voting (60\%), exhibit high TPRs alongside high FPRs, failing to meet the criteria for an ideal combination strategy.

Comparison with various multiple testing schemes in Table~\ref{tab6} reveals that the TPR levels of Bonferroni, BY, and BH surpass the method proposed in this study, indicating a less conservative control of TPR due to the DOS-Storey correction method. Despite a TPR value of 94.91\%, the proposed method shows a minor impact on the TPR target level. Among these schemes, the BY algorithm achieves the highest TPR but with significantly higher FPR values, while Bonferroni, BH, and the proposed method exhibit TPR values of 95.07\%, 94.96\%, and 94.91\% respectively, with the proposed method demonstrating the lowest FPR value. Specifically, compared to the optimal combination scheme, the proposed method reduces the FPR value from 3.83\% to 3.31\%, effectively balancing the trade-off between TPR and FPR. Overall, the DSDE-KNN detector maintains a low FPR while minimally impacting the TPR, showcasing significant advantages in the analysis.

\subsection{Larger model library}

This section examines the impact of expanding the model library on the FPR.
A comparative experiment was conducted in this section before and after enlarging the model library for OoD detection. The experiments detailed the use of CIFAR10 as the in-distribution dataset and the KNN method as the OoD detector with a parameter value of $K=50$. In Section 4.2, eight additional pre-trained models, including WideResNet~\cite{zagoruyko2016wide}, ResNeXt50~\cite{he2016identity}, ResNeXt101~\cite{he2016identity}, ResNeXt152~\cite{he2016identity}, PreResNet18~\cite{xie2017aggregated}, PreResNet34~\cite{xie2017aggregated}, PreResNet50~\cite{xie2017aggregated},  PreResNet101~\cite{xie2017aggregated}, SwinV2-B256, SwinV2-B384, and SwinV2-L256, were incorporated into the model library originally used for CIFAR10 experiments. The original set of models in Section 4.2 was expanded to encompass a total of 18 pre-trained models. The hyperparameters for Algorithm \ref{algorithm1} were specifically defined as $c_{m} = 2/15$ and $\beta = 1$.

The experimental findings denote "Small" for the detector utilizing the original model library in Section 4.2 and "Large" for the detector post the expansion of the model library. The detailed experimental results are presented in Table~\ref{tab5}.

\begin{table}
\caption{DSDE-KNN Detector based on the Original Model Library vs. DSDE-KNN Detector based on the Expanded Model Library.}\label{tab5}
\resizebox{\textwidth}{!}{
\begin{tabular}{|l|l|l|l|l|l|l|}
\hline
 \makecell[c]{Method/(\%)} & \makecell[c]{TPR \\ ~}  & \makecell[c]{SVHN \\(FPR/AUC)} & \makecell[c]{LSUN \\ (FPR/AUC)} & \makecell[c]{iSUN \\ (FPR/AUC)} & \makecell[c]{DTD \\ (FPR/AUC)} & \makecell[c]{Places365 \\ (FPR/AUC)} \\
\hline
DSDE-KNN(Small) & 94.91 & 1.92/99.46 & {\bf 1.37/99.63} & 4.89/98.77 & 0.12/99.88 & 8.26/98.17\\
DSDE-KNN(Large) & 94.85 & {\bf 0.04/99.92} & 1.66/99.56 & {\bf 4.13/98.92} & {\bf 0.00/100.00} & {\bf 0.00/100.00}\\
\hline
\end{tabular}}
\end{table}

The detector with an expanded model library showcases a remarkable 97.92\% reduction in False Positive Rate (FPR) values on the SVHN dataset and a notable 15.54\% decrease in FPR values on the iSUN dataset, as outlined in Table~\ref{tab5}. Notably, the detector with the extended model library demonstrated exceptional performance on the challenging DTD dataset and Places365 dataset, reducing the FPR value to 0 and ensuring 100\% accuracy in identifying OoD samples. The flawless recognition on the DTD dataset and Places365 dataset further support the theory proposed by Xue et al. However, the performance of the detector based on the expanded model library showed some instability on the LSUN datasets. 

\section{Conclusion}

This study aims to enhance post-hoc OoD detection performance by utilizing various pre-trained models in the model library. Initially, the research explores different combinations of single-model detectors and proposes a method to systematically integrate the detection decision statistics of multiple detectors using a multiple testing framework. Utilizing the Storey FDR correction method within this framework, the study investigates the estimation of true null hypothesis proportions. To improve the accuracy of this estimation, the paper introduces the DOS-Storey proportion estimator and establishes a novel OoD detection method, named DOS-Storey-based Detector Ensemble (DSDE). 
The efficacy of our proposed method was validated in comprehensive experiments. Additionally, the KNN method enhanced detector showcased outstanding performance in OoD detection, reducing the false positive rate from 20.74\% to 3.31\% compared to the optimal baseline method.

%
%
%
\bibliographystyle{splncs04}
\bibliography{samplepaper}

\end{document}